%% file: beyond-fisher.tex
\documentclass[11pt]{article}

\usepackage{amsmath,amsbsy,amsfonts,amssymb,amsthm,dsfont,fullpage,units,psfrag}
\usepackage{graphicx}
\usepackage{algorithm,algorithmic,mathtools}
\usepackage{color,cases}

\usepackage{tikz}
\usetikzlibrary{calc, shapes, backgrounds, arrows, fit, calc}
\usepackage{subfigure}
\usepackage{hyperref}
\usepackage[round]{natbib}
\usepackage{appendix}

 \newcommand{\IGNORE}[1]{}

\newcommand\E{\mathbb{E}}
\newcommand\R{\mathbb{R}}

\def\Pc{{\cal S}}

\newcommand\tl{\tilde}

\input{macros}

\author{Majid Janzamin\footnote{University of California, Irvine. Email: mjanzami@uci.edu} \and Hanie Sedghi\footnote{University of Southern California. Email: hsedghi@usc.edu} \and Anima Anandkumar\footnote{University of California, Irvine. Email: a.anandkumar@uci.edu}}

\title{Score Function Features for Discriminative Learning: \\ Matrix and Tensor Framework}

\begin{document}
\maketitle

\begin{abstract}Feature learning forms the cornerstone for tackling challenging learning problems in domains such as speech, computer vision and natural language processing. In this paper, we consider a novel class of matrix and tensor-valued features, which can be pre-trained using unlabeled samples. We present efficient algorithms for extracting discriminative information, given these pre-trained features and labeled samples for any related task. Our class of   features are based on higher-order score functions, which capture local variations in the probability density function of the input. We establish a theoretical framework to characterize the nature of discriminative information that can be extracted from score-function features, when used in conjunction with labeled samples.  We employ efficient spectral decomposition algorithms (on matrices and tensors) for extracting discriminative components. The advantage of employing tensor-valued features is that we can extract richer discriminative information in the form of an overcomplete representations. Thus, we present a novel framework for employing   generative models of the input for discriminative learning. \end{abstract}


\input{intro}

\input{related}

\input{problem-formulation}

\input{results-anima}

\input{tensor-decomposition}

\input{Discussion}

\input{theorems}

\subsubsection*{Acknowledgements}
M. Janzamin thanks Rina Panigrahy for useful discussions.
M. Janzamin is supported by NSF Award CCF-1219234. H. Sedghi is supported by ONR Award N00014-14-1-0665. A. Anandkumar is supported in part by Microsoft Faculty Fellowship, NSF Career award CCF-$1254106$, NSF Award CCF-$1219234$, ARO YIP Award W$911$NF-$13$-$1$-$0084$ and ONR Award N00014-14-1-0665.

\renewcommand{\appendixpagename}{Appendix}

\appendixpage

\appendix


\section{Proof of Theorems} \label{appendix:proof_higher}

\bprfof{Theorem~\ref{steinslemma_par}}
Let us denote by $\nabla_\theta h(\theta_0)$ the derivative of function $h(\theta)$ w.r.t.\ $\theta$ evaluated at point $\theta_0$. We have
\begin{align*}
\int \nabla_\theta \bigl( G(x;\theta_0)p(x;\theta_0) \bigr) dx
&= \int \nabla_\theta G(x;\theta_0) \cdot p(x;\theta_0) dx + \int G(x;\theta_0) \otimes \nabla_\theta p(x;\theta_0) dx \\
&= \int \nabla_\theta G(x;\theta_0) \cdot p(x;\theta_0) dx + \int G(x;\theta_0) \otimes \nabla_\theta \log  p(x;\theta_0) \cdot p(x;\theta_0) dx \\
&= \Ebb[\nabla_{\theta}  G(x;\theta_0)] + \Ebb[G(x;\theta_0) \otimes \nabla_{\theta} \log p(x;\theta_0)],
\end{align*}
where the first step is concluded from product rule. On the other hand, we have
\[
\int \nabla_\theta \bigl( G(x;\theta_0)p(x;\theta_0) \bigr) dx
= \nabla_\theta \int G(x;\theta_0)p(x;\theta_0) dx
= \nabla_\theta c_g = 0,
\]
where the second and first regularity conditions are respectively exploited in the above steps. Combining the above two inequalities, the result is proved.
\eprfof

\bprfof{Theorem~\ref{thm:steins_higher}}
The proof is done by iteratively applying the recursion formula of score functions in~\eqref{eqn:diffoperator_recursion} and Stein's identity in Lemma~\ref{steinslemma}. First, we provide the first order analysis as follows:
\begin{align*}
\Ebb \left[ G(x) \otimes \Pc_m(x) \right]
& {\overset{(e_1)}{=}} - \Ebb \left[ G(x) \otimes \Pc_{m-1}(x) \otimes \nabla_x \log p(x) \right] - \Ebb \left[ G(x) \otimes \nabla_x \Pc_{m-1}(x) \right] \\
& {\overset{(e_2)}{=}} \ \Ebb \left[ \nabla_x \left( G(x) \otimes \Pc_{m-1}(x) \right) \right] - \Ebb \left[ G(x) \otimes \nabla_x \Pc_{m-1}(x) \right] \\
& {\overset{(e_3)}{=}} \ \Ebb \left[ \nabla_x G(x) \otimes \Pc_{m-1}(x) \right]^{\langle \pi \rangle} + \Ebb \left[ G(x) \otimes \nabla_x \Pc_{m-1}(x) \right]  - \Ebb \left[ G(x) \otimes \nabla_x \Pc_{m-1}(x) \right] \\
& = \ \Ebb \left[ \nabla_x G(x) \otimes \Pc_{m-1}(x) \right]^{\langle \pi \rangle},
\end{align*}
where recursion formula~\eqref{eqn:diffoperator_recursion} is used in equality $(e_1)$,  equality $(e_2)$ is concluded by applying Stein's identity in Lemma~\ref{steinslemma} for which we also used the regularity condition that all the entries of $G(x) \otimes \Pc_{m-1}(x) \otimes p(x)$ go to zero on the boundaries of support of $p(x)$. Finally, the product rule in Lemma~\ref{lem:prodrule} is exploited in $(e_3)$ with appropriate permutation vector $\pi$ to put the differentiating variable in the last mode of the resulting tensor.

By iteratively applying above steps, the result is proved. Note that the permutation in the final step does not affect on the tensor $\nabla_x^{(m)} G(x)$ which is symmetric along the involved modes in the permutation.
\eprfof

\subsection{Auxiliary lemmas}

%

\begin{lemma} [Product rule for gradient] \label{lem:prodrule}
Consider $F(x)$ and $G(x)$ as tensor-valued functions
\begin{align*}
&F(x) : \R^n \rightarrow\bigotimes^{p_1} \R^n, \\
&G(x) : \R^n \rightarrow\bigotimes^{p_2} \R^n.
\end{align*}
Then, we have
\[
\nabla (F(x) \otimes G(x)) = (\nabla F(x) \otimes G(x))^{\langle \pi \rangle} + F \otimes \nabla G(x),
\]
for permutation vector $\pi = [1,2,\dotsc,p_1,p_1+2,p_1+3,\dotsc,p_1+p_2+1,p_1+1]$.
\end{lemma}

\bprf
The lemma is basically the product rule for derivative with the additional transposition applied to the first term. The necessity for transposition is argued as follows.

Note that for tensor-valued function $F(x)$, the gradient $\nabla F(x)$ is defined in~\eqref{eqn:derivativedef} such that the last mode of the gradient tensor $\nabla F(x)$ corresponds to the entries of derivation argument or variable $x$. This is immediate to see that the transposition applied to first term $\nabla F(x) \otimes G(x)$ is required to comply with this convention. This transposition enforced by the specified permutation vector $\pi$ puts the last mode of $\nabla F(x)$ (mode number $p_1+1$) to the last mode of whole tensor $\nabla F(x) \otimes G(x)$. Note that such transposition is not required for the other term $F \otimes \nabla G(x)$ since the last mode of $\nabla G(x)$ is already the last mode of $F \otimes \nabla G(x)$ as well.
\eprf

We also prove the explicit form of score functions in~\eqref{eqn:diffoperator} as follows.
\begin{equation*}
\Pc_m(x)=(-1)^m \frac{\nabla_x^{(m)} p(x)}{p(x)}.
\end{equation*}

\bprfof{explicit score function form in~\eqref{eqn:diffoperator}}
The result is proved by induction. It is easy to verify that the basis of induction holds for $m=0,1$. Now we argue the inductive step assuming that the result holds for $m-1$ and showing that it also holds for $m$. Substituting the induction assumption in the recursive form of $\Pc_m(x)$ defined in~\eqref{eqn:diffoperator_recursion}, we have
\begin{align*}
\Pc_m(x) & = (-1)^m \frac{\nabla_x^{(m-1)} p(x)}{p(x)} \otimes \nabla_x \log p(x) + (-1)^m \nabla_x \left( \frac{\nabla_x^{(m-1)} p(x)}{p(x)} \right) \\
& = (-1)^m \frac{\nabla_x^{(m-1)} p(x) \otimes \nabla_x p(x)}{p(x)^2} + (-1)^m \frac{p(x) \nabla_x^{(m)} p(x) - \nabla_x^{(m-1)} p(x) \otimes \nabla_x p(x) }{p(x)^2} \\
& = (-1)^m \frac{\nabla_x^{(m)} p(x)}{p(x)},
\end{align*}
where the quotient rule for derivative is used in the second equality.
\eprfof


\end{document}

%% file: macros.tex
\def\tl{\tilde}

 \def\0{{\bf 0}}

\def\qed{\hfill\hbox{${\vcenter{\vbox{
    \hrule height 0.4pt\hbox{\vrule width 0.4pt height 6pt
    \kern5pt\vrule width 0.4pt}\hrule height 0.4pt}}}$}}

\definecolor{myred}{rgb}{0.3,0.0,0.7}
\definecolor{dkg}{rgb}{0.1,0.7,0.2}
\definecolor{dkb}{rgb}{0.0,0.2,0.8}

 \def\hu{\hat{u}}

\def\Hc{{\cal H}}

\def\Nc{{\cal N}}

\def\Ebb{{\mathbb E}}

\def\Rbb{{\mathbb R}}

\def\Zbb{{\mathbb Z}}

\newcommand{\bprfof}{\begin{proof_of}}
\newcommand{\eprfof}{\end{proof_of}}
\newcommand{\bprf}{\begin{myproof}}
\newcommand{\eprf}{\end{myproof}}
\newcommand{\bp}{\begin{psfrags}}
\newcommand{\ep}{\end{psfrags}}
\newcommand{\bl}{\begin{lemma}}
\newcommand{\el}{\end{lemma}}
\newcommand{\bt}{\begin{theorem}}
\newcommand{\et}{\end{theorem}}
\newcommand{\bc}{\begin{center}}
\newcommand{\ec}{\end{center}}
\newcommand{\bi}{\begin{itemize}}
\newcommand{\ei}{\end{itemize}}
\newcommand{\ben}{\begin{enumerate}}
\newcommand{\een}{\end{enumerate}}
\newcommand{\bd}{\begin{definition}}
\newcommand{\ed}{\end{definition}}
\def\beq{\begin{equation}}
\def\eeq{\end{equation}\noindent}
\def\beqn{\begin{eqnarray}}
\def\eeqn{\end{eqnarray} \noindent}
\def\beqnn{  \begin{eqnarray*}}
\def\eeqnn{\end{eqnarray*}  \noindent}
\def\bcase{  \begin{numcases}}
\def\ecase{\end{numcases}   \noindent}
\def\bsbcase{  \begin{subnumcases}}
\def\esbcase{\end{subnumcases}   \noindent}

\newtheorem{theorem}{Theorem}

\newtheorem{lemma}[theorem]{Lemma}

\newtheorem{definition}{Definition}

\newenvironment{myproof}{\noindent{\bf Proof:} \hspace*{1em}}{
    \hspace*{\fill} $\Box$ }
\newenvironment{proof_of}[1]{\noindent {\bf Proof of #1: }}{\hspace*{\fill} $\Box$ }

\newcommand{\matplottc}[1]{               
        \unitlength .45truein
        \begin{center}
        \includegraphics{#1.ps}
        \end{picture}
        \end{center}
}

\def\psfancypar#1#2{\begingroup\def\par{\endgraf\endgroup\lineskiplimit=0pt}
               \setbox2=\hbox{\large\sc #2}
               \newdimen\tmpht \tmpht \ht2 \advance\tmpht by \baselineskip
               \font\hhuge=Times-Bold at \tmpht
               \setbox1=\hbox{{\hhuge #1}}
               \count7=\tmpht \count8=\ht1
               \divide\count8 by 1000 \divide\count7 by \count8
               \tmpht=.001\tmpht\multiply\tmpht by \count7
               \font\hhuge=Times-Bold at \tmpht
               \setbox1=\hbox{{\hhuge #1}}
               \noindent
                \hangindent1.05\wd1
               \hangafter=-2 {\hskip-\hangindent
               \lower1\ht1\hbox{\raise1.0\ht2\copy1}%
                \kern-0\wd1}\copy2\lineskiplimit=-1000pt}

\def\Kout{\setbox1=\hbox{\Huge\bf K}\hbox to
1.05\wd1{\hspace{.05\wd1}
}}
\def\Sout{\setbox1=\hbox{\Huge\bf S}\hbox to 1.05\wd1{\hspace{.05\wd1}
}}

\newcommand{\torestate}[3]{%
\expandafter \def \csname BBRESTATE #2 \endcsname{#3}
\theoremstyle{plain}
\newtheorem{BBRESTATETHMNUM#2}[theorem]{#1}
\begin{BBRESTATETHMNUM#2}\label{#2}\csname BBRESTATE #2 \endcsname   \end{BBRESTATETHMNUM#2}
\newtheorem*{BBRESTATETHMNONNUM#2}{{#1}~\ref{#2}}
}

\newcommand{\restate}[1]{\begin{BBRESTATETHMNONNUM#1}[Restated] \csname BBRESTATE #1 \endcsname
\end{BBRESTATETHMNONNUM#1}}


%% file: intro.tex
\section{Introduction}

Having good features or representations of the input data  is critical to achieving good performance in challenging machine learning tasks in domains such as speech, computer vision and natural language processing~\citep{bengio2013representation}. Traditionally, feature  engineering   relied on carefully hand-crafted features,  tailored towards a specific task: a laborious and a time-consuming process.
Instead, the recent trend has been to automatically learn good features through various frameworks such as deep learning~\citep{bengio2013representation}, sparse coding~\citep{raina2007self}, independent component analysis (ICA)~\citep{le2011ica}, Fisher kernels~\citep{jaakkola1999exploiting}, and so on. These approaches are unsupervised and can thus exploit the vast amounts of unlabeled samples, typically present in these domains.

A good feature representation  incorporates important prior knowledge about the input, typically through a probabilistic model. In almost every conceivable scenario, the probabilistic model needs to incorporate latent variables to fit the input data. These latent factors can be important explanatory variables for classification  tasks associated with the input. Thus, incorporating generative models of the input can hugely boost the performance of discriminative tasks.

Many approaches to feature learning  focus   on unsupervised  learning, as described above.
The hypothesis behind employing unsupervised  learning is  that the input distribution is related to the associative model between the input and the label of a given task, which is reasonable to expect in most scenarios.  When the distribution of the unlabeled samples, employed for feature learning, is the same as the labeled ones, we have the framework of   {\em semi-supervised} learning. A more general framework, is the so-called {\em self-taught} learning, where the distribution of unlabeled samples is different, but related to the labeled ones~\citep{raina2007self}. Variants of these frameworks include transfer learning, domain adaptation and multi-task learning~\citep{bengio2012deep}, and involve labeled datasets for related tasks.  These frameworks have been of extensive interest to the machine learning community, mainly  due to the scarcity of labeled samples for many challenging tasks. For instance,  in   computer vision, we have a huge corpus of unlabeled images, but a more limited set of labeled ones. In natural language processing, it is extremely laborious to annotate the text with syntactic and semantic parses, but we have access to unlimited amounts of unlabeled text.

%


It has been postulated that humans mostly learn in an unsupervised manner~\citep{raina2007self}, gathering ``common-sense'' or ``general-purpose'' knowledge, without worrying about any specific goals. Indeed, when faced with a specific task, humans  can quickly and easily extract  relevant   information  from the accrued  general-purpose knowledge.  Can we design machines with similar capabilities? Can we design algorithms which  succinctly summarize  information in  unlabeled samples as  general-purpose features? When given a specific task, can we efficiently extract relevant  information from general-purpose features? Can we provide  theoretical guarantees for such algorithms? These are indeed challenging questions, and we   provide some concrete answers  in this paper.


\subsection{Summary of Results}

In this paper, we consider a class of matrix and tensor-valued ``general-purpose'' features,  pre-trained  using unlabeled samples. We assume that the labels are not present at the time of feature learning. When presented with labeled samples, we leverage these pre-trained features to  extract discriminative information using efficient spectral decomposition algorithms. As a main contribution, we provide theoretical guarantees on the nature of discriminative information that can be extracted with our approach.


We consider the class of features based on higher-order score functions of the input, which involve higher-order derivatives of the   probability density function (pdf). These functions    capture  ``local manifold structure'' of the pdf. While the first-order score function is a vector (assuming a vector input), the higher-order functions are matrices and tensors, and thus capture richer information about the input distribution.
Having access to these matrix and tensor-valued features allows to extract better discriminative information, and we characterize its precise nature in this work.

Given  score-function features and  labeled samples,   we extract discriminative information  based on the   method of moments. We construct cross-moments involving the labels and the input score features. Our main theoretical result is  that these moments are equal to the expected derivatives of the label, as a function of the input or some model parameters. In other words, these moments capture  variations of the label function, and are therefore informative for discriminative tasks.

We   employ  spectral decomposition algorithms  to find succinct representations of the moment matrices/tensors. These algorithms are fast  and embarrassingly parallel. See ~\citep{AnandkumarEtal:tensor12,JanzaminEtal:Altmin14,JanzaminEtal:Altmin14-2} for details, where we have developed and analyzed efficient tensor decomposition algorithms (along with our collaborators). The advantage of the tensor methods is that they do not suffer from spurious local optima, compared to typical non-convex problems such as expectation maximization or backpropagation in neural networks. Moreover,
we can construct overcomplete representations for tensors, where the number of  components in the representation can exceed the data dimensionality.
It has been  argued that having overcomplete   representations  is crucial to getting good classification performance~\citep{coates2011analysis}.
Thus, we can leverage the latest advances in spectral methods for efficient extraction of  discriminative information from moment  tensors.

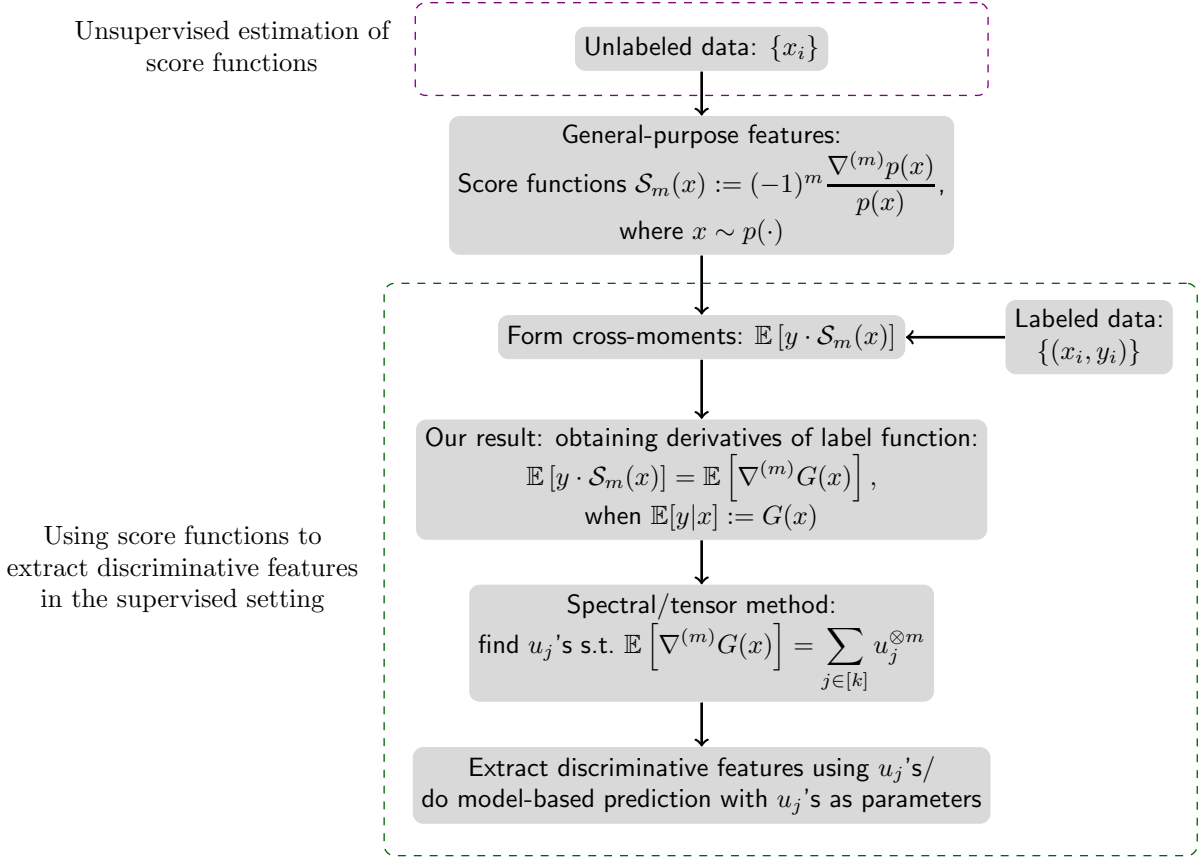
\begin{figure}[t]
\bc
\begin{tikzpicture}
[
scale=1,
   nodestyle/.style={fill = gray!30, shape = rectangle, rounded corners, minimum width = 2cm, font = \sffamily},
]
\small
\matrix [column sep=2mm,row sep=6mm] {
\node[nodestyle](a){Unlabeled data: $\{x_i\}$}; & \\
\node[nodestyle, align=center](b){General-purpose features: \\ Score functions $\Pc_m(x) := (-1)^m \dfrac{\nabla^{(m)} p(x)}{p(x)}$, \\ where $x \sim p(\cdot)$}; & \\
\node[nodestyle, align=center](c){Form cross-moments: $\displaystyle \Ebb \left[y \cdot \Pc_m(x) \right]$}; & \node[nodestyle, align=center](c2){Labeled data: \\ $\{(x_i,y_i)\}$}; \\
\node[nodestyle, align=center](d){Our result: obtaining derivatives of  label function: \\ $\displaystyle \Ebb \left[y \cdot \Pc_m(x) \right] = \Ebb \left[ \nabla^{(m)} G(x) \right],$ \\ when $\Ebb[y|x]:=G(x)$ }; & \\
\node[nodestyle, align=center](e){Spectral/tensor method: \\ find $u_j$'s s.t.\ $\displaystyle\Ebb \left[ \nabla^{(m)} G(x) \right]= \sum_{j \in [k]} u_j^{\otimes m}$}; & \\
\node[nodestyle, align=center](f){Extract discriminative features using $u_j$'s/ \\ do model-based prediction with $u_j$'s as parameters}; & \\
};
\draw [->, line width = 1pt] (a) to (b);
\draw [->, line width = 1pt] (b) to (c);
\draw [->, line width = 1pt] (c) to (d);
\draw [->, line width = 1pt] (d) to (e);
\draw [->, line width = 1pt] (e) to (f);
\draw [->, line width = 1pt] (c2) to (c);

\node [draw, dashed, rounded corners, violet, line width=0.5pt,
	fit = {(a) ($(a.east)+(20mm,0)$) ($(a.west)-(20mm,0)$) ($(a.north)+(0,2mm)$) ($(a.south)-(0,2mm)$)},
	label=left:{\begin{tabular}{c} Unsupervised estimation of\\ score functions \end{tabular}}
	] {};

\node [draw, dashed, rounded corners, green!40!black, line width=0.5pt,
	fit = {(c) (c2) (d) (e) (f) ($(c2.east)+(2mm,0)$) ($(d.west)-(3mm,0)$) ($(c.north)+(0,3mm)$) ($(f.south)-(0,3mm)$)},
	label=left:{\begin{tabular}{c} Using score functions to \\ extract
discriminative features \\ in the supervised setting \end{tabular}}
	] {};

\end{tikzpicture}
\ec
\caption{Overview of the proposed framework of using the general-purpose features to generate discriminative features through spectral methods.}\label{fig:overview}
\end{figure}

In our framework, the  label can be a scalar,  a vector, a matrix or even a tensor, and it can either be continuous or discrete. We can therefore handle a variety of regression and classification settings such as multi-task, multi-class,   and structured prediction problems.
Thus we present a unified and an efficient end-to-end framework for extracting discriminative information from pre-trained features. An overview of the entire framework is presented in Figure~\ref{fig:overview}.

We now provide some important observations below.

\paragraph{Are the expected label function derivatives informative? }
Our analysis characterizes the discriminative information we can extract from score function features. As described above, we prove that  the cross-moments between the label and the score function features are equal to the expected derivative of the label as a function of the input or model parameters. But  when are these expected label derivatives informative? Indeed, in trivial cases, where the derivatives of the label function vanish over the  support of the input distribution, these moments carry no information. However, such cases are pathological, since then, either there is  no variation   in  the label function or the input distribution is nearly degenerate. Another possibility is that  a certain derivative vanishes, when averaged over the input distribution, even though it is not zero everywhere. If this is the case, then the next derivative cannot be averaged out to zero, and will thus carry information about the variations of the label function. Thus, in practical scenarios, the cross-moments contain useful discriminative information. In fact,  for many discriminative models which are challenging to learn, such as multi-layer neural networks and mixtures of classifiers, we establish that these moments have an intimate relationship with the parameters of the discriminative model in subsequent works~\citep{Sedghi:SparseNet,Sedghi:mixture}. Spectral decomposition of the moments provably recovers the model parameters. These are the first results for guaranteed  learning of many challenging discriminative latent variable models.

\paragraph{Contrasting with previous approaches: }We now contrast our approach to previous approaches for incorporating generative models in discriminative tasks. Typically, these approaches directly feed the pre-trained features   to a classifier. For example, in the Fisher kernel framework, the Fisher score features are fed to a kernel classifier~\citep{jaakkola1999exploiting}. The reasoning behind this is that the  features obtained from unsupervised learning have information  about all the classes, and the task of finding class-specific differences in the learnt representation is left to the  classifier. However, in practice, this may not be the case, and a common complaint is that these generative features are not discriminative for the task at hand. Previous solutions have prescribed joint training discriminative features using labeled samples, in conjunction with unlabeled samples~\citep{mairal2009supervised,maaten2011learning,wang2013robust}. However, the resulting optimization problems are complex and expensive to run, may not converge to good solutions, and have to be re-trained for each new task. We present an alternative approach to extract discriminative features using efficient spectral decomposition algorithms on   moment  matrices and tensors. These methods are light weight and fast,   and we theoretically quantify the nature of discriminative features they can extract. These discriminative features can then be fed into the classification pipeline.  Thus, the advantage of our approach is that we can quickly generate discriminative features for new classification tasks without going through the laborious process of re-training for   new features.

We now contrast our approach with previous moment-based approaches for discriminative learning, which   consider moments between the label and  raw input, e.g.~\citep{karampatziakis2014discriminative}. Such methods have no theoretical guarantees. In contrast, we construct cross-moments between the label and the score function features.   We show that using score function features   is crucial to mining discriminative information with provable guarantees.

\paragraph{Extension to self-taught learning: }
We have so far described our framework under the semi-supervised setting, where the unlabeled and labeled samples have the same input distribution. We can also handle the framework of self-taught learning, where the two  distributions are related but may not be the same.  We prescribe some simple pre-processing to transfer the parameters and to re-estimate the score function  features  for the input of the labeled data set. Such parameter transfer frameworks have been considered before, e.g. \citep{raina2007self}, except here we present a general latent-variable framework and  focus on transferring parameters for computing score functions, since we require them for subsequent operations. Our framework can also  be applied to   scenarios  where we  have different input sources with different distributions, but the classification task is the same, and thus, the associative model between the label and the input is fixed. Consider for instance,   crowdsourcing applications, where the same task is presented to different groups of individuals. In our approach, we can then construct different score function features for different input sources and the different cross-moments provide  information about the variations in the label function, averaged over different input distributions. We can thus leverage the diversity of different input sources for improved performance on  common tasks. Thus, our approach is applicable in many challenging practical scenarios. 

\subsection{Overview of our framework}

In   this section, we elaborate on the end-to-end framework presented in Figure~\ref{fig:overview}.

\paragraph{Background:} The problem of supervised learning consists of learning a predictor, given labeled training samples $\{(x_i, y_i)\}$ with input   $x_i$ and corresponding  label $y_i$. Classical frameworks such as SVMs are purely discriminative since they make no distributional assumptions. However, when labeled data is limited and classification tasks are challenging,  incorporating distributional information can improve performance.  In an associative  model-based framework, we   posit a conditional distribution for the label given the input $p(y|x)$. However, learning this model is challenging, since maximum-likelihood estimation of $p(y|x)$ is non-convex and NP-hard to solve in general, especially if it involves hidden variables (e.g., associative mixtures, multi-layer neural networks). In addition, incorporating a generative model for input $x$ often leads to improved   discriminative performance.

\paragraph{Label-function derivatives are discriminative:}
Our main focus in this work is to extract useful information about $p(y|x)$ without attempting to learn it in its entirety. In particular, we extract  information about the local variations of conditional distribution $p(y|x)$, as the  input $x$  (or some model parameter) is changed. For the classification setting, it suffices to consider\footnote{In the classification setting, powers of $y$, e.g., $y^2$ contain no additional information, and hence, all the information of the associative model is in $\Ebb[y|x]:=G(x)$. However, in the regression setting, we can compute additional functions, e.g., $\Ebb[\nabla^{(m)} H(x)]$, where $\Ebb[y^2 | x] := H(x)$. Our approach can also compute these derivatives.}   $\Ebb[y|x]:=G(x)$. In this paper, we present mechanisms to estimate its expected higher order derivatives\footnote{Note that since we are computing the expected derivatives, we also assume a distribution for  the input $x$.}
\beq \label{eqn:derivative-intro} \Ebb[\nabla_x^{(m)} G(x)], \ m \geq 1,\eeq
where $\nabla^{(m)}_x$ denotes the $m$-th order derivative operator w.r.t.\ variable $x$.
By having access to expected derivatives of the label function   $G(x)$ in \eqref{eqn:derivative-intro}, we   gain an understanding of how the label $y$ varies  as we   change the input $x$ locally, which is valuable discriminative information.


\paragraph{Score functions yield  label-function derivatives:}
One of the main contributions of this paper is to obtain these expected derivatives  in \eqref{eqn:derivative-intro} using features denoted by  $\Pc_m (x)$, for $m \geq 1$ (learnt from unlabeled samples) and the labeled data. In particular, we form the cross-moment between the label $y$ and the features $\Pc_m(x)$, and show that they yield the derivatives as\footnote{We drop subscript $x$ in the derivative operator $\nabla^{(m)}_x$ saying $\nabla^{(m)}$ when there is no ambiguity.}
\beq \Ebb[ y \cdot \Pc_m(x)] =   \Ebb[\nabla^{(m)} G(x)], \quad \mbox{when }\Ebb[y|x]:= G(x).\label{eqn:yield}\eeq

We establish a simple form  for features  $\Pc_m(x)$,  based on the derivatives of the probability density function $p(\cdot)$ of the input $x$ as
\beq \label{eqn:highorderintro}\Pc_m(x) = (-1)^m\frac{\nabla^{(m)} p(x)}{p(x)}, \quad \mbox{when } x \sim p(\cdot). \eeq
In fact, we show that the feature $\Pc_m(x)$ defined above is a function of higher order score functions $\nabla_x^{(n)} \log p(x)$ with $n \leq m$, and we derive an  explicit relationship between them. This is basically why we also call these features as (higher order) score functions. Note that the features $\Pc_m(x)$ can be learnt using unlabeled samples, and we term them  as general-purpose features since they can be applied to any labeled dataset, once they are estimated. Note the features $\Pc_m(x)$ can be   vectors, matrices or  tensors, depending on $m$, for multi-variate $x$. The choice of order $m$ depends on the particular setup: a higher $m$ yields more information (in the form of higher order derivatives) but requires more samples to compute the empirical moments accurately.

We then extend the framework to parametric setting, where we obtain derivatives $\Ebb[\nabla^{(m)}_\theta G(x;\theta)]$ with respect to some model parameter $\theta$ when $\Ebb[y|x;\theta]:=G(x;\theta)$. These are obtained using general-purpose features denoted by $\Pc_m(x;\theta)$ which  is a function of higher order Fisher score functions $\nabla_\theta^{(n)} \log p(x;\theta)$ with $n \leq m$. Note that by  using the parametric framework   we can now incorporate discrete input $x$, while this is not possible with the previous framework.

\paragraph{Spectral decomposition of derivative matrices/tensors:}
Having obtained the derivatives $\Ebb[\nabla^{(m)} G(x)]$  (which are  matrices or tensors), we then find efficient representations  using spectral/tensor decomposition  methods. In particular, we find vectors $u_j$ such that \beq\label{eqn:decomp} \Ebb[\nabla^{(m)} G(x)] =\sum_{j \in [k]} \overbrace{u_j\otimes u_j\otimes \cdots \otimes u_j}^{m \mbox{ times}},\eeq where $\otimes$ refers to the tensor product notation. Note that since the higher order derivative is a symmetric matrix/tensor, the decomposition is also symmetric. Thus, we decompose the matrix/tensor at hand into sum of rank-$1$ components, and in the matrix case, this reduces to computing the SVD.  In the case of a tensor, the above decomposition is termed as CP decomposition~\citep{Kruskal:77}. In a series of works~\citep{AnandkumarEtal:tensor12,JanzaminEtal:Altmin14,JanzaminEtal:Altmin14-2}, we have presented efficient algorithms for obtaining \eqref{eqn:decomp}, and analyzed their performance in detail. 


The matrix/tensor in hand is decomposed into a sum of $k$ rank-1 components. Unlike matrices,  for tensors, the  rank parameter $k$ can be larger than the dimension. Therefore, the decomposition problems falls  in to two different regimes. One is the undercomplete regime: where $k$ is less than the dimension, and the overcomplete one, where it is not. The undercomplete regime  leads to  dimensionality reduction, while the overcomplete regime results in richer representation. 

Once we obtain components $u_j$, we then have several options to perform further processing. We can extract discriminative features such as $\sigma(u_j^\top x)$, using some non-linear function $\sigma(\cdot)$, as performed in some of the earlier works, e.g.,~\citep{karampatziakis2014discriminative}.
Alternatively, we can perform model-based prediction and incorporate $u_j$'s as parameters of a discriminative model.  In a subsequent paper, 
we show that $u_j$'s correspond to significant   parameters of many challenging discriminative models such as multi-layer feedforward neural networks and mixture of classifiers, under the {\em realizable} setting.

\paragraph{Extension to self-taught learning: }
The results presented so far assume the semi-supervised setting, where the unlabeled samples $\{\tilde{x}_i\}$ used to estimate the score functions are drawn from the same distributions as the  input $\{x_i\}$ of the labeled samples $\{(x_i, y_i)\}$. We present simple mechanisms to extend to the self-taught setting, where the distributions of $\{\tilde{x}_i\}$ and $\{x_i\}$ are related, but not the same. We assume latent-variable models for $\tilde{x}$ and $x$, e.g., sparse coding, independent component analysis (ICA), mixture models, restricted Boltzmann machine (RBM), and so on. We assume that the conditional distributions $p(\tilde{x}| \tilde{h})$ and $p(x|h)$, given the corresponding latent variables $\tilde{h}$ and $h$ are the same. This is reasonable since the unlabeled samples $\{\tilde{x}_i\}$ are usually ``rich'' enough to cover all the elements. For example,  in the sparse coding setting, we assume that all the dictionary elements can be learnt through $\{\tilde{x}_i\}$, which is assumed in a number of previous works, e.g~\citep{raina2007self,zhang2008flexible}. Under this assumption, estimating the score function for new samples $\{x_i\}$ is relatively straightforward, since we can transfer the estimated conditional distribution $p(\tl{x}|\tl{h})$ (using unlabeled samples $\{\tl{x}_i\}$) as the estimation of $p(x|h)$, and we can re-estimate the marginal distribution $p(h)$ easily. Thus, the use of  score functions allows for easy transfer of information under the self-taught framework. The rest of the steps can proceed as before.

%% file: related.tex
\subsection{Related Work}

Due to limitation of labeled samples in many domains such as computer vision and natural language processing, the frameworks of domain adaptation, semi-supervised,   transfer and multi-task learning have been popular in domains such as   NLP~\citep{blitzer2006domain}, computer vision~\citep{quattoni2007learning,yang2007cross,hoffman2013efficient}, and so on. We now list the various approaches below.

\paragraph{Non-probabilistic approaches: }Semi-supervised learning has been extensively studied via non-probabilistic approaches. Most works  attempt to assign labels to the unlabeled samples, e.g.~\citep{ando2005framework}, either through bootstrapping~\citep{yarowsky1995unsupervised,blum1998combining}, or by learning good functional structures~\citep{szummer2002partially,ando2005framework}. Related to semi-supervised learning is the problem of   domain adaptation, where the source domain has  labeled datasets on which classifiers have been trained, e.g. ~\citep{ben2006analysis,huang2006correcting,NIPS2008_3550,blitzer2009zero,ben2010theory,gong2013connecting}, and there may or may not be labeled samples in the target domain.  The main difference is that in this paper, we consider the source domain to have only unlabeled samples, and we pre-train general-purpose features, which are not tied to any specific task.

A number of recent works have investigated transfer learning using deep neural networks, e.g.~\citep{bengio2012deep,socher2013zero,zeiler2013visualizing,sermanet2013overfeat,donahue2013decaf,YosinskiCBL14} and obtain state-of-art performance on various tasks.



\paragraph{Probabilistic approaches (Fisher kernels): }A number of works explore probabilistic approaches to semi-supervised and transfer learning, where they learn a generative model on the input and use the features from the model for discriminative tasks. Fisher kernels fall into this category, where the Fisher score is pre-trained using unlabeled samples, and then used for discriminative tasks through a kernel classifier~\citep{jaakkola1999exploiting}. Our paper proposes higher order extensions of the Fisher score, which yield matrix and tensor score features, and we argue that they are much more informative for discriminative learning, since they yield higher order derivatives of the label function. Moreover, we provide a different mechanism to utilize the score features: instead of directly feeding the score features to the classifier, we form cross-moments between the score features and the labels, and extract discriminative features through spectral decomposition. These discriminative features can then be used in the standard classification frameworks. This allows us to overcome a common complaint that the pre-trained features, by themselves may not be discriminative for a particular task. Instead there have been attempts to construct discriminative features from generative models using labeled samples~\citep{maaten2011learning,mairal2009supervised,wang2013robust}. However, this is time-consuming since the discriminative features  need to be re-trained for every new   task.

\paragraph{Probabilistic approaches (latent representations): }A number of works learn latent representations to obtain higher level features for classification tasks. Popular models include sparse coding
~\citep{raina2007self}, independent component analysis (ICA)~\citep{le2011ica}, and restricted Boltzmann machines (RBM)~\citep{swersky2011autoencoders}. It has been argued that having overcomplete latent representations, where the latent dimensionality exceeds the observed dimensionality, is crucial to getting good classification performance~\citep{coates2011analysis}. We also note here that the authors and others have been involved in developing guaranteed and efficient algorithms for unsupervised learning of latent representations such as mixture models, ICA~\citep{AnandkumarEtal:tensor12,JanzaminEtal:Altmin14-2},   sparse coding~\citep{AnandkumarEtal:COLT14,arora2014new}, and deep representations~\citep{arora2013provable}. There have been various other probabilistic frameworks for information transfer.~\citet{raina2006constructing} consider learning priors  over parameters in one domain and using it in the other domain in the Bayesian setting.~\citet{mccallum2006multi}   argue that incorporating generative models for $p(x|y)$ acts as a regularizer and leads to improved performance.

~\citet{raina2007self} introduce the framework of self-taught learning, where the distribution of unlabeled samples is related but not the same as the input for labeled samples. They employ a sparse coding model for the input, and assume that both the datasets share the same dictionary elements, and only the distribution of the coefficients  which combine the dictionary elements are different. They learn the dictionary from the unlabeled samples, and then  use it to decode the input of the labeled  samples. The decoded  dictionary coefficients are the features which are fed into SVM for classification. In this paper, we provide an alternative framework for learning features in a self-taught framework by transferring parameters for score function estimation in the new domain.

\paragraph{Probabilistic approaches (score matching): }We now review the score matching approaches for learning probabilistic models. These methods estimate parameters using score matching criteria rather than a likelihood-based one. Since we utilize score function features, it makes sense to estimate parameters based on the score matching criterion.~\citet{hyvarinen2005estimation} introduce the criterion of minimizing the Fisher divergence, which is the expected square loss between the model score function and the data score function. Note that the score function $\nabla_x \log p(x)$ does not involve the partition function, which is typically intractable to compute, and is thus tractable in scenarios where the likelihood cannot be computed.~\citet{lyu2009interpretation} further provide a nice theoretical result  that the score matching criterion is more robust than maximum likelihood in the presence of noise. ~\citet{swersky2011autoencoders} apply the score matching framework for learning RBMs, and extract features for classification, which show superior performance compared to auto-encoders.


\paragraph{Probabilistic approaches (regularized auto-encoders): }Another class of approaches for feature learning are the class of regularized auto-encoders. An auto-encoder maps the input to a code through an encoder function, and then maps back using a decoder function.  The training criterion is to minimize the reconstruction loss along with a regularization penalty. They have been employed for pre-training neural networks. See~\citep{bengio2013representation} for a review.~\citet{vincent2011connection} established that a special case of denoising auto-encoder reduces to a score matching criterion for an appropriately chosen energy function. ~\citet{alain2012regularized} establish that  the denoising auto-encoders estimate the score function (of first and second order), in the limit as the noise variance goes to zero. In this paper, we argue that the score function are the appropriate features to learn for transferring information to various related tasks. Thus, we can employ auto-encoders for estimating the score functions.

\paragraph{Stein's identity: }Our results establishing that the higher order score functions in \eqref{eqn:highorderintro} yield derivatives of the label function in \eqref{eqn:yield}   is novel.   The special case of the first derivative   reduces to  Stein's identity in statistics~\citep{stein1986approximate, ley2013parametric}, which   is essentially obtained through integration by parts. We construct  higher order score functions in a recursive manner, and then show that it reduces to the simple form in \eqref{eqn:highorderintro}.

\paragraph{Orthogonal polynomials:} For the special case of Gaussian input $x \sim \Nc(0,I)$,  we show that the score functions  in \eqref{eqn:highorderintro}  reduce to the familiar multivariate {\em Hermite} polynomials, which are orthogonal polynomials for the Gaussian distribution~\citep{MultiHermite:Grad1949,MultiHermite:Holmquist96}. However, for general distributions, the score functions  need not be polynomials.



%% file: problem-formulation.tex
\section{Problem Formulation} \label{sec:problem-formulation}

In this section, we first review different learning settings; in particular semi-supervised and self-taught learning settings which we consider in this paper. Then, we state the main assumptions to establish our theoretical guarantees.

\subsection{Learning settings and assumptions}

First, we describe different learning settings and clarify the differences between them by giving some image classification examples, although the framework is general and applicable to other domains as well. 



\paragraph{Semi-supervised learning:}
In the semi-supervised setting, we have both labeled samples $\{(x_i,y_i)\}$ and unlabeled samples $\{\tilde{x}_i\}$ in the training set. For instance, consider a set of images   containing cats and dogs, where a fraction of them are labeled with the binary output $y_i$ specifying if the image contains cat or dog. The main assumption is that the input in the labeled and unlabeled datasets have the same distribution.

\paragraph{Multi-task learning:}
In the multi-task setting, we have labeled samples $\{(x_i,y_i)\}$ and $\{(x_i,\tl{y}_i)\}$ where the same set of inputs $x_i$ are labeled for two different tasks.  For instance, consider a set of images containing cats, dogs, monkeys and humans where the first task is to label them as \{human, not human\}, while the other task is to label them as  \{cat, not cat\}.


\paragraph{Transfer learning:}
In  transfer learning, we want to exploit the labeled information of one task to perform other related tasks. This is also known as knowledge transfer since the goal is to transfer the knowledge gained in analyzing one task to another. Concretely, we have access to labeled samples $\{(x_i,y_i)\}$ and $\{(\tl{x}_i,\tl{y}_i)\}$ of two related tasks. For instance, imagine a set of images $\{(x_i,y_i)\}$ containing cats and dogs, another set of images $\{(\tl{x}_i,\tl{y}_i)\}$ containing monkeys and humans, each of them with the corresponding labels. The goal is to  use the information in source labeled data $\{(\tl{x}_i,\tl{y}_i)\}$ for classifying new samples of target data $x_i$.

\paragraph{Self-taught learning:}
In  self-taught learning, we further assume that the related dataset $\{\tl{x}_i\}$ in the transfer learning setting does not have labels. Concretely, we have labeled samples of the original task as $\{(x_i,y_i)\}$, and other unlabeled data  $\{\tl{x}_i\}$ from a related distribution. For instance, consider a set of images $\{(x_i,y_i)\}$ containing cats and dogs with labels, and assume we  have lots of unlabeled images $\{\tl{x}_i\}$ which can be any type of images, say downloaded from internet.

\paragraph{ }
In this paper, we focus on semi-supervised and self-taught learning settings, and other related learning frameworks mentioned above are also treated in these two settings, i.e. we consider first training score function features from input, without using labels, and then use the labels in conjunction with score function features.
 
We first give general assumptions we use in both semi-supervised and self-taught settings. Then, we state additional assumptions for the self-taught learning framework.

\paragraph{Probabilistic input $x$:}
We assume a generative model on input $x$ where it is randomly drawn from some continuous distribution $p(x)$ satisfying mild regularity condition\footnote{The exact form of regularity conditions are provided in Theorem~\ref{thm:steins_higher}.}. It is known that incorporating such generative assumption on $x$ usually results in better performance on discriminative tasks.

\paragraph{Probabilistic output $y$:}
We further assume  a probabilistic model on output (label) $y$ where it is randomly drawn according to some distribution $p(y|x)$ given input $x$, satisfying mild regularity conditions\footnote{The exact form of regularity conditions are provided in Theorem~\ref{thm:steins_higher}.}. In our framework, the output (label) can be scalar, vector, matrix or even tensor, and it can be continuous or discrete, and we can handle a variety of regression and classification settings such as multi-class, multi-label, and structured prediction problems. 

\paragraph{Assumptions under the self-taught learning framework:} We now state the assumptions that tie the distribution of unlabeled samples with labeled ones.   We consider latent-variable models for $\tilde{x}$ and $x$, e.g., sparse coding, independent component analysis, mixture models, restricted Boltzmann machine, and so on. We assume that the conditional distributions $p(\tilde{x}| \tilde{h})$ and $p(x|h)$, given the corresponding latent variables $\tilde{h}$ and $h$, are the same. This is reasonable since the unlabeled samples $\{\tilde{x}_i\}$ are usually ``rich'' enough to cover all the elements. For example,  in the sparse coding setting, we assume that all the dictionary elements can be learnt through $\{\tilde{x}_i\}$, which is assumed in a number of previous works, e.g~\citep{raina2007self,zhang2008flexible}.
In particular, in the image classification task mentioned earlier, consider a set of images $\{x_i\}$ containing cats and dogs, and assume we also have lots of unlabeled images $\{\tl{x}_i\}$, which can be any type of images, say downloaded from internet. Under the sparse coding model, the observed images are the result of a sparse combination of dictionary elements. The coefficients for combining the dictionary elements correspond to hidden variables $h$ and $\tilde{h}$ for the two datasets. It is reasonable to expect that the two datasets share the same dictionary elements, i.e., once the coefficients are fixed, it is reasonable to assume that the conditional probability of drawing the observed pixels in images is the same for both labeled images (including only cats and dogs) and unlabeled images (including all random images). But  the marginal probability of the coefficients, denoted by $p(h)$ and $p(\tl{h})$, will be  different, since they represent two  different data sets.  In Section~\ref{sec:score-self}, we show that this assumption leads to simple mechanisms to transfer knowledge about the score function to the new dataset.

\subsection{A snapshot of our approach}

We now succinctly explain our general framework and state how the above assumptions are involved in our setting. In general, semi-supervised learning considers access to both labeled and unlabeled samples in the training data set. When the labeled data is limited and the learning task mostly relies on the unlabeled data (which is the case in many applications), the task is more challenging, and assuming distributional assumptions as above can improve the performance of learning task. 

Note that maximum likelihood estimation of $p(y|x)$ is non-convex and NP-hard in general, and our goal in this work is to extract useful information from $p(y|x)$ without entirely recovering it. We extract   information about the local variations of conditional distribution $p(y|x)$, when input $x$ is changed. In particular, for the classification task, we extract useful information from derivatives (including local variation information) of the first order conditional moment of output $y$ given input $x$ denoted as\,\footnote{In the classification setting, powers of $y$, e.g., $y^2$ contain no additional information, and hence, all the information of the associative model is in the first order conditional moment $\Ebb[y|x]:=G(x)$. However, in the regression setting, we can involve higher order conditional moments, e.g., $\Ebb[y^2 | x] := H(x)$. Our approach can also compute the derivatives of these higher order conditional moments.}
$$\E[y|x] := G(x).$$
More concretely, we provide mechanisms to compute $\E[\nabla_x^{(m)} G(x) ]$, where $\nabla^{(m)}_x$ denotes the $m$-th order derivative operator w.r.t.\ variable $x$. We usually limit to a small $m$, e.g., $m=3$. 

Note that in computing  $\E[\nabla_x^{(m)} G(x) ]$,   we also apply the expectation over input $x$, and thus, the generative model of $x$ comes into the picture. This derivative is a vector/matrix/tensor, depending on $m$. Finally, we decompose this derivative matrix/tensor to rank-1 components to obtain discriminative features.

Now the main question is how to estimate the derivatives $\E[\nabla^{(m)} G(x) ]$. One of our main contributions in this work is to  show  that the {\em score functions} yield such label-function derivatives.
For $m=1$, it is known that the (first order) score function yields the derivative as
$$ - \Ebb[y \cdot \nabla \log p(x) ] = \Ebb[\nabla G(x)],  \quad \mbox{when }\Ebb[y|x]:= G(x), $$
where $-\nabla \log p(x)$ is the (usual first order) score function. More generally, we introduce ($m$-th order) score functions denoted by $\Pc_m(x)$ which also yield the desired derivatives as
$$ \Ebb[y \cdot \Pc_m(x) ] = \Ebb[\nabla^{(m)} G(x)]. $$
The estimation of score functions $\Pc_m(x)$ is performed in an unsupervised manner using unlabeled samples $\{x_i\}$; see Section~\ref{sec:score-estimation} for the details.

\paragraph{Computing cross-moment between label $y$ and score function $\Pc_m(x)$:} After estimating the score function, we form the cross moment $\Ebb[y \cdot \Pc_m(x) ]$ between labels $y$ and (higher order) score functions $\Pc_m(x)$ using labeled data.
Here, we assume that we can compute the {\em exact} form of these moments. Perturbation analysis of the computed empirical moment depends on the setting of the probabilistic models on input $x$ and output $y$ which is the investigation direction in  the subsequent works applying the proposed framework in this paper to specific learning tasks.

%% file: results-anima.tex
\section{Score Functions Yield Label Function Derivatives: Informal Results}

In this section, we provide one of our main contributions in this paper,  which is showing that higher order score functions    yield {\em differential operators}. 
Here, we   present   informal statements of the main result, along with with detailed discussions, while the formal lemmas and theorems are stated in Section~\ref{sec:theorems}. 

\subsection{First order score functions}
We first review the existing results on score functions and their properties as yielding first order differential operators. The score function is the derivative of the logarithm of density function. The derivative is w.r.t.\ either variable $x$ or the parameters of the distribution. The latter one is usually called {\em Fisher score} in the literature.  We provide the properties of score functions as yielding differential operators in both cases.

\paragraph{Stein identity:}
We start with Stein's identity (or Stein's lemma) which is the building block of our work. The original version of Stein's lemma is for the Gaussian distribution. 
For a standard random Gaussian vector $x \sim \Nc(0,I_{d_x})$, it states that for all  functions\footnote{We consider general tensor valued functions $G(x):\mathbb{R}^{d_x} \rightarrow \bigotimes^r \mathbb{R}^{d_y}$. For details on tensor notation, see Section~\ref{sec:notation}.} $G(x)$ satisfying mild regularity conditions, we have~\citep{stein1972} 
\begin{equation} \label{eqn:Steinbasic}
\Ebb[G(x) \otimes x]= \Ebb[\nabla_x G(x)],
\end{equation}
where $\otimes$ denotes the tensor product (note that if $G(x)$ is a vector (or a scalar), the notation $G(x) \otimes x$ is equivalent to $G(x) x^\top$), and $\nabla_x$ denotes the usual gradient operator w.r.t.\ variable $x$. For details on tensor and gradient notations, see Section~\ref{sec:notation}.

The above result for the Gaussian distribution can be generalized to 
other random distributions as follows. For a random vector $x \in \R^{d_x}$, let $p(x)$ and $\nabla_x \log p(x)$ respectively denote the joint density function and the corresponding {\em score function}. Then, under some mild regularity conditions, for all   functions $G(x)$, we have
\begin{align}
\label{eq:stein}
\Ebb[G(x) \otimes \nabla_x \log p(x)]=-\Ebb[\nabla_x G(x)].
\end{align}
See Lemma~\ref{steinslemma} for a formal statement of this result including description of regularity conditions. Note that for the Gaussian random vector $x\sim \Nc(0,I_{d_x})$ with joint density function $p(x) = \frac{1}{\left(\sqrt{2\pi}\right)^{d_x}} e^{-\|x\|^2/2}$, the score function is $\nabla_x \log p(x) = -x$, and the above equality reduces to the special case in~\eqref{eqn:Steinbasic}.

\paragraph{Parametric Stein identity:}A 
parametric approach to Stein's lemma is introduced in~\citep{ley2013parametric}, which we review   here. 
We first define some   notations. Let $\Theta$ denote the set of parameters such that for $\theta \in \Theta$, $p(x;\theta)$ be a valid $\theta$-parametric probability density function. In addition, consider any $\theta_0 \in \Theta$ for which specific regularity conditions for a class of functions $G(x;\theta)$ hold over a neighborhood of $\theta_0$. See Definition~\ref{def_par} for a detailed description of the regularity conditions, which basically allows us to  change the order of derivative w.r.t.\ to $\theta$ and integration on $x$. 
We are now ready to state the informal parametric Stein's identity as follows.


For a random vector $x \in \R^{d_x}$, let $p(x;\theta)$ and $\nabla_\theta \log p(x;\theta)$ respectively denote the joint $\theta$-parametric density function and the corresponding {\em parametric score function}. Then, for all  functions $G(x;\theta)$ satisfying the above (mild) regularity conditions, we have
\begin{equation}
\label{eq:steinpar}
\Ebb[G(x;\theta) \otimes \nabla_{\theta} \log p(x;\theta)] = -\Ebb[\nabla_{\theta}  G(x;\theta)] \quad \textnormal{at} \ \theta=\theta_0.
\end{equation}
See Theorem~\ref{steinslemma_par} for a formal statement of this result including description of regularity conditions.

\paragraph{Contrasting the above Stein identities:}
We provide Stein's identity and the parametric form in~\eqref{eq:stein} and \eqref{eq:steinpar}, respectively. The two identities mainly differ in taking the derivative w.r.t.\ either the variable $x$ or the parameter $\theta$. We now provide an example assuming the mean vector as the parameter of the distribution to elaborate the parametric result and contrast it with the original version, where we also see how the two forms are closely related in this case.

Consider the random Gaussian vector $x \in \R^d$ with mean parameter $\mu$ and known identity covariance matrix. Hence,
$$p(x;\mu) = \frac{1}{\left(\sqrt{2\pi}\right)^{d}} e^{-(x-\mu)^\top (x-\mu)/2}$$
denotes the corresponding joint parametric density function with mean parameter $\theta = \mu$. Thus, the parametric score function is $\nabla_{\mu} \log p(x;\mu) = x-\mu$
and applying parametric Stein's identity in \eqref{eq:steinpar} to functions of the form $G(x;\mu) = G_0(x - \mu)$ leads to
$$\Ebb[G_0(x-\mu_0) \otimes (x-\mu_0)] = -\Ebb[\nabla_{\mu} G_0(x-\mu) \vert_{\mu=\mu_0}].$$
Setting $\mu_0=0$, this identity is the same as the original Stein's identity in~\eqref{eqn:Steinbasic} since $\nabla_{\mu} G_0(x-\mu) = - \nabla_x G_0(x-\mu)$. 

Note that this relation is true for any distribution and not just Gaussian, i.e., for the joint  parametric density function $p(x;\mu)$ with mean parameter $\mu$, we have the Stein identities from \eqref{eq:stein} and \eqref{eq:steinpar} respectively as
\begin{align*}
\Ebb[G_0(x-\mu_0) \otimes \nabla_x \log p(x;\mu)] & =-\Ebb[\nabla_x G_0(x-\mu_0)], \\
\Ebb[G_0(x-\mu) \otimes \nabla_{\mu} \log p(x;\mu)] &= -\Ebb[\nabla_{\mu}  G_0(x-\mu)] \quad \textnormal{at} \ \mu=\mu_0,
\end{align*}
which are the same since $\nabla_{\mu} G_0(x-\mu) = - \nabla_x G_0(x-\mu)$ and $\nabla_x \log p(x;\mu) = -\nabla_{\mu} \log p(x;\mu)$.


\subsection{Higher order score functions}

The first order score functions and their properties as yielding differential operators are reviewed in the previous section. Such differential property is called the  Stein's  identity. In this section, we generalize such differential properties to higher orders by introducing higher order score functions as matrices and tensors.

\subsubsection{Our contribution: higher order extensions to Stein's identities}

Let $p(x)$ denote the joint probability density function of random vector $x \in \R^d$. We denote $\Pc_m(x)$ as the $m$-th order score function,  which we establish is given by\,\footnote{Since $\Pc_m(x)$ is related to $m$-th order derivative of the function $p(x)$ with input vector $x$, it represents a tensor of order $m$, i.e., $\Pc_m \in \bigotimes^m \R^d$. For details on tensor notation, see Section~\ref{sec:notation}.}
\beq\label{eqn:highorder}
\Pc_m(x)=(-1)^m \frac{\nabla_x^{(m)} p(x)}{p(x)},
\eeq
where $\nabla_x^{(m)}$ denotes the $m$-th order derivative operator w.r.t.\ variable $x$.
Note that the first order score function $\Pc_1(x) = -\nabla_x \log p(x)$ is the same as the score function  in~\eqref{eq:stein}. Furthermore, we show that $\Pc_m(x)$ is equivalently constructed from the recursive formula
 \begin{equation} \label{eqn:diffoperator_recursion_informal}
\Pc_m(x) = - \Pc_{m-1}(x) \otimes \nabla_x \log p(x) - \nabla_x \Pc_{m-1}(x),
\end{equation}
with $\Pc_0(x)=1$. Here $\otimes$ denotes the tensor product; for details on tensor notation, see Section~\ref{sec:notation}.  Thus, $\Pc_m(x)$ is related to higher order score functions $\nabla_x^{(n)} \log p(x)$ with $n \leq m$, which is the reason we also call $\Pc_m(x)$'s as higher order score functions.
These functions $\Pc_m(x)$  enable us to generalize the Stein's identity in~\eqref{eq:stein} to higher order derivatives, i.e., they yield higher order differential operators.

\begin{theorem}[Higher order differential operators, {\em informal statement}] \label{thm:steins_higher_informal}
For random vector $x$, let $p(x)$ and $\Pc_m(x)$ respectively denote the joint density function and the corresponding $m$-th order score function in \eqref{eqn:highorder}. Then, under some mild regularity conditions, for all functions $G(x)$, we have
\[
\Ebb \left[ G(x) \otimes \Pc_m(x) \right] = \Ebb \left[ \nabla^{(m)}_x G(x) \right],
\]
where $\nabla_x^{(m)}$ denotes the $m$-th order derivative operator w.r.t.\ variable $x$.
\end{theorem}

See Theorem~\ref{thm:steins_higher} for a formal statement of this result including description of regularity conditions.

\paragraph{Comparison with orthogonal polynomials:}
In the case of standard multivariate Gaussian distribution as $x \sim \Nc(0,I_d)$, the score functions defined in~\eqref{eqn:highorder} turn out to be multivariate {\em Hermite} polynomials~\citep{MultiHermite:Grad1949,MultiHermite:Holmquist96}  $\Hc_m(x)$  defined as
$$
\Hc_{m}(x) := (-1)^m  \frac{\nabla^{(m)}_x p(x)}{p(x)}, \quad p(x) = \frac{1}{\left(\sqrt{2 \pi}\right)^{d}} e^{-\|x\|^2/2}.
$$    It is also worth mentioning that the Hermite polynomials satisfy the orthogonality property as~\citep[Theorem 5.1]{MultiHermite:Holmquist96}
\begin{align*}
 \Ebb \left[ \Hc_m(x) \otimes \Hc_{m'}(x) \right] =
\left\{\begin{array}{ll}
m! I^{\otimes m}, & m=m', \\
0, & \operatorname{otherwise},
\end{array} \right.
\end{align*}where the   expectation is over the standard multivariate Gaussian distribution. The application of higher order Hermite polynomials as yielding differential operators has been known before~\citep[equation (37)]{GoldsteinReinert2005}, but applications have mostly involved  scalar variable $x \in \R$.

Thus, the proposed higher order score functions $\Pc_m(x)$ in \eqref{eqn:highorder} coincides with the orthogonal Hermite polynomials $\Hc_m(x)$ in  case of multivariate Gaussian distribution.  However, this is not necessarily the case for other distributions; for instance, it is convenient to see that the Laguerre polynomials which are orthogonal w.r.t.\ Gamma distribution are different from the score functions proposed in~\eqref{eqn:highorder}, although the Laguerre polynomials have a differential operator interpretation too; see \citet{GoldsteinReinert2005} for the details.
Note that the proposed score functions need not be polynomial functions in general.

\subsubsection{Parametric higher order Stein identities}
In this section, we provide the generalization of first order parametric differential property in~\eqref{eq:steinpar} to higher orders. In order to do this, we  introduce the parametric form of higher order score functions in \eqref{eqn:highorder}.
Let $\Pc_m(x;\theta) $ be the $m$-th order parametric score function  given by
\beq\label{eqn:highorderparam}
\Pc_m(x;\theta)=(-1)^m \frac{\nabla_{\theta}^{(m)} p(x;\theta)}{p(x;\theta)}.
\eeq
Similar to the previous section, we can construct $\Pc_m(x;\theta)$ as a function of higher order Fisher score functions $\nabla^{(n)}_\theta \log p(x;\theta)$, $n \leq m$, as
\begin{equation} \label{eqn:diffoperator_recursion_par_informal}
\Pc_m(x;\theta) := - \Pc_{m-1}(x;\theta) \otimes \nabla_\theta \log p(x;\theta) - \nabla_\theta \Pc_{m-1}(x;\theta),
\end{equation}
with $\Pc_0(x;\theta)=1$.


Note that the first order parametric score function $\Pc_1(x;\theta) = -\nabla_\theta \log p(x;\theta)$ is the same as Fisher score function exploited in~\eqref{eq:steinpar}.
These higher order score functions enable us to generalize the parametric Stein's identity in~\eqref{eq:steinpar} to higher orders as follows.

\begin{theorem}[Higher order parametric differential operators, {\em informal statement}] \label{thm:steins_higher_par_informal}
For random vector $x \in \R^{d_x}$, let $p(x;\theta)$ and $\Pc_m(x;\theta)$ respectively denote the joint $\theta$-parametric density function and the corresponding $m$-th order score function in~\eqref{eqn:highorderparam}.
Then, for all   functions $G(x;\theta)$ satisfying the regularity conditions, we have
\[
\Ebb[G(x;\theta) \otimes \Pc_m(x;\theta) ] = \Ebb[\nabla_{\theta}^{(m)} G(x;\theta)] \quad \textnormal{at} \ \theta=\theta_0.
\]
\end{theorem}

See Theorem~\ref{thm:steins_higher_par} for a formal statement of this result including description of regularity conditions.

The advantage of this parametric form is it can be applied to both discrete and continuous random variables.

%% file: tensor-decomposition.tex
\section{Spectral Decomposition Algorithm} \label{sec:tensor}

As part of the framework we introduced in Figure~\ref{fig:overview}, we need a spectral/tensor method to decompose the higher order derivative tensor $\Ebb[\nabla^{(m)} G(x)]$ to its rank-1 components denoted by $u_j$. Let us first consider the case that the derivative tensor is a matrix\,\footnote{For instance, it happens when the label function $y$ is a scalar, and $m=2$ for vector input $x$. Then, $\Ebb[\nabla^{(2)} G(x)]$ is a matrix (second order tensor).}. Then the problem of decomposing this matrix to the rank-1 components reduces to the usual Principle Component Analysis (PCA), where the rank-1 directions are the eigenvectors   of the matrix.


More generally, we can form higher order derivatives ($m>2$) of the label function $G(x)$ and extract more information from their decomposition. The higher order derivatives are represented as {\em tensors} which can be seen as multi-dimensional arrays. There exist different tensor decomposition frameworks, but the most popular one is the CP decomposition where a (symmetric) rank-$k$ tensor $T  \in \Rbb^{d \times d \times d}$ is written as the sum of $k$ rank-$1$ tensors\,\footnote{The decomposition for an asymmetric tensor is similarly defined as $T = \sum_{j\in [k]} u_j \otimes v_j \otimes w_j, \ u_j,v_j,w_j \in \Rbb^d$.}
\begin{equation} \label{eqn:tensor-decomposition}
T = \sum_{j\in [k]} u_j \otimes u_j \otimes u_j, \quad u_j \in \Rbb^d.
\end{equation}
Here notation $\otimes$  represents the tensor (outer) product; see Section~\ref{sec:notation} for a detailed discussion on the tensor notations.

We now state a tensor decomposition algorithm for computing decomposition forms in~\eqref{eqn:tensor-decomposition}.
The Algorithm~\ref{algo:Power method form} is considered by~\citet{JanzaminEtal:Altmin14} where the generalization to higher order tensors can be similarly introduced. The main step in~\eqref{eqn:power update} performs {\em power iteration}\,\footnote{This is the generalization of matrix power iteration to $3$rd order tensors.}; see~\eqref{eqn:rank-1 update} for the multilinear form definition.
After running the algorithm for all different initialization vectors, the clustering process from~\citet{JanzaminEtal:Altmin14} ensures that the best converged vectors are returned as the estimates of true components $u_j$. Detailed analysis of the tensor decomposition algorithm and its convergence properties are provided by~\citet{JanzaminEtal:Altmin14}. We briefly summarize   the initialization and convergence guarantees of the algorithm below.

\begin{algorithm}[t]
\caption{Tensor decomposition via tensor power iteration~\citep{JanzaminEtal:Altmin14}}
\label{algo:Power method form}
\begin{algorithmic}
\REQUIRE 1) Rank-$k$ tensor $T = \sum_{j \in [k]} u_j \otimes u_j \otimes u_j \in \Rbb^{d \times d \times d}$, 2) $L$ initialization vectors $\hat{u}^{(1)}_\tau$, $\tau \in [L]$, 3) number of iterations $N$.
\FOR{$\tau=1$ \TO $L$}
\FOR{$t=1$ \TO $N$}
\STATE Tensor power updates (see \eqref{eqn:rank-1 update} for the definition of the multilinear form):
\begin{equation} \label{eqn:power update}
\hu_\tau^{(t+1)} = \frac{T \left( I, \hu_\tau^{(t)}, \hu_\tau^{(t)} \right)}{\left\| T \left( I, \hu_\tau^{(t)}, \hu_\tau^{(t)} \right) \right\|}, \quad
\end{equation}
\ENDFOR
\ENDFOR
\RETURN the cluster centers of set $\left\{ \hu_\tau^{(N+1)} : \tau \in [L] \right\}$ (by Procedure~\ref{alg:cluster}) as estimates $u_j$.
\end{algorithmic}
\end{algorithm}

\floatname{algorithm}{Procedure}
\begin{algorithm}[t]
\caption{Clustering process~\citep{JanzaminEtal:Altmin14}}
\label{alg:cluster}
\begin{algorithmic}
\REQUIRE Tensor $T \in \Rbb^{d \times d \times d}$, set 
$S := \left\{ \hu_\tau^{(N+1)}:\tau \in [L] \right\}$, parameter $\nu$.
\WHILE{$S$ is not empty} 
\STATE Choose $u \in S$ which maximizes $|T(u,u,u)|$.
\STATE Do $N$ more iterations of power updates in \eqref{eqn:power update} starting from $u$.
\STATE Let the output of iterations denoted by $\tilde{u}$ be the center of a cluster.
\STATE Remove all the $u \in S$ with $|\langle u, \tilde{u} \rangle| > \nu/2$.
\ENDWHILE
\RETURN the cluster centers.
\end{algorithmic}
\end{algorithm}

\paragraph{Initialization:}Since tensor decomposition is a non-convex problem, different initialization lead to different solutions.~\citet{JanzaminEtal:Altmin14} introduce two initialization methods for the above algorithm. One is random initialization and the other is a SVD-based technique, where the convergence analysis is provided  for the latter one.

\paragraph{Convergence guarantees:}
Tensor power iteration is one of the key algorithms for decomposing rank-$k$ tensor $T$ in~\eqref{eqn:tensor-decomposition} into its rank-1 components $u_j$'s.~\citet{ZG01} provide the convergence analysis of tensor power iteration in the orthogonal setting where the tensor components $u_j$'s are orthogonal to each other, and~\citet{AnandkumarEtal:tensor12} analyze the robustness of this algorithm to noise.

Note that  the rank-$k$ tensor decomposition can still be unique even if the rank-1 components are not orthogonal (unlike the matrix case).~\citet{JanzaminEtal:Altmin14} provide local and global convergence guarantees for Algorithm~\ref{algo:Power method form} in the non-orthogonal and overcomplete, (where the tensor rank $k$ is larger than the dimension $d$) settings. The main assumption in their analysis is the incoherence property which imposes soft-orthogonality conditions on the components $u_j$'s; see~\citet{JanzaminEtal:Altmin14} for  details.

\paragraph{Whitening (orthogonalization):}
In the non-orthogonal and undercomplete (where the tensor rank $k$ is smaller than the dimension $d$), instead of direct application of tensor power iteration for tensor decomposition as in Algorithm~\ref{algo:Power method form}, we first orthogonalize the tensor and then apply the tensor power iteration, which requires different perturbation analysis; see for instance \citet{SongEtal:NonparametricTensorDecomp}. In the orthogonalization step also known as whitening, the tensor modes are multiplied by whitening matrix such that the resulting tensor has an orthogonal decomposition.

%% file: Discussion.tex
\section{Unsupervised Estimation of Score Functions} \label{sec:score-estimation}

In this section, we discuss further on the   score function  and its estimation. First, we discuss the form of score function for  exponential family, and  for models with latent variables. Next, we review the frameworks which estimate the score function and discuss the connection with auto-encoders. Note that these frameworks can be also extended to learning score functions of a nonlinear transformation of the data. 




\subsection{Score function for exponential family}
The score function expression proposed in Equation~\eqref{eqn:highorder}
can be further simplified when the random vector $x$ belongs to the exponential family distributions where we have $p(x;\theta) \propto \exp(-E(x;\theta))$. Here, $E(x;\theta)$ is known as the energy function. Then, we have

%
\begin{align*}
\Pc_m(x)&=(-1)^m \sum_{\alpha_1, \dotsc, \alpha_t} \nabla^{(\alpha_1)}_x E(x,\theta) \otimes \nabla^{(\alpha_2)}_x E(x,\theta) \otimes \cdots \otimes \nabla^{(\alpha_t)}_x E(x,\theta), \\
& \text{where } \{\alpha_i \in \Zbb^+, \ i \in [t]: \sum_{i=1}^t \alpha_i=m\}.
\end{align*}
Thus, in case of the exponential family, the higher order score functions are compositions of the derivatives of the energy function.

\subsection{Score function for Latent Variable Models}
For a latent variable model $p(x,h;\theta)$, let $h$ denote the vector of latent variables and $x$ the vector of observed ones. It is well known that the (first order)  Fisher score function of $x$ is the marginalized of the joint Fisher score~\citep{tsuda2002marginalized}
\begin{align*}
\nabla_{\theta} \log p (x;\hat{\theta})=\sum_h p(h|x;\hat{\theta}) \nabla_{\theta} \log p(x,h;\hat{\theta}).
\end{align*}  Given this marginalized form, \citet{tsuda2002marginalized} also show that Fisher kernel is a special case of marginalized kernels (where the joint kernel over both observed and latent variables is marginalized over the posterior distribution).

For the special case of Gaussian mixtures, we can simplify the above general form in the following manner.
Let $x = Ah + z$, where $z$ has multivariate standard normal distribution for simplicity. In this case, the score function $\nabla_x \log p(x)$ is equal to $x-A \Ebb[h|x]$,
 where $E[h|x]$ is the posterior estimation of the mean. This means that the mean vectors are weighted with posterior estimation of the mean, i.e., centering based on contribution of each mixture. Note that if $h$ were observed, we would be centering based on the mean of that component. But since $h$ is hidden, we center based on the posterior distribution of $h$.
 %


The higher order score functions can be readily calculated as 
\[ \Pc_m = (-1)^m \frac{\sum_h \nabla^{(m)} p(x,h)}{\sum_h p(x,h)}.\]

\subsection{Efficient Estimation of the Score Function}
There are various efficient methods for computing the score function. In deep learning, the framework of auto-encoders attempts to find encoding and decoding functions which minimize the reconstruction error under noise (the so-called denoising auto-encoders or DAE). This is an unsupervised framework involving only unlabeled samples.~\citet{alain2012regularized} argue that the DAE   approximately learns the score function of the input, as the noise variance goes to zero. Moreover, they also describe ways to estimate the second order score function.


\paragraph{Score matching:}
The framework of score matching is popular for parameter estimation  in probabilistic models~\citep{hyvarinen2005estimation, swersky2011autoencoders}, where the criterion is to fit parameters is based on matching the data score function. We now review the score matching framework and analysis in~\citet{lyu2009interpretation}.
Let\,\footnote{For the sake of notation simplicity, we also refer to $p(x)$ and $q_\theta(x)$ as $p$ and $q_\theta$ respectively, i.e., dropping the dependence on $x$.} $p(x)$ denote the pdf  of  $x$, and the goal is to find a parametric probabilistic model $q_\theta(x)$ with model parameter $\theta$ that best matches $p(x)$. \citet{lyu2009interpretation} formulate the score matching framework introduced by~\citep{hyvarinen2005estimation} as minimizing the Fisher divergence between two distributions $p(x)$ and $q_\theta(x)$, defined as
\begin{align*}
D_F(p \Vert q_{\theta}) := \int_{x} p(x) \left\| \frac{\nabla_x p(x)}{p(x)} - \frac{\nabla_x q_{\theta}(x)}{q_{\theta}(x)} \right\|^2 dx.
\end{align*}
Note that $\frac{\nabla_x p(x)}{p(x)} = \nabla_x \log p(x)$ is  the first order score function (up to sign).
\citet{lyu2009interpretation} also show that the Fisher divergence can be equivalently written as
\begin{align*}
D_F(p \Vert q_{\theta})=\int_{x} p(x) \left( \left\| \nabla \log p(x) \right\|^2 + \left\| \nabla \log q_\theta(x) \right\|^2 + 2 \bigtriangleup \log q_\theta(x) \right).
\end{align*}
where $\bigtriangleup$ denotes the Laplacian operator $\bigtriangleup := \sum_{i \in [d]} \frac{\partial^2}{\partial x_i^2}$. Then, they also provide a nice interpretation of Fisher divergence relating that to the usual KL (Kulback-Leibler) divergence $D_{KL}(p \Vert q_{\theta}) := \int_{x} p(x) \log \frac{p(x)}{q_\theta(x)} dx$ in the sense of robustness to Gaussian noise as follows. Note that this also gives the relation between score matching and maximum-likelihood (ML) estimation since ML is achieved by minimizing the KL-divergence.
\begin{lemma}[\citealt{lyu2009interpretation}]
Let⃗ $y = x+ \sqrt{t} w$, for $t \geq 0$ and $w$ a zero-mean white Gaussian vector. Denote $\tl{p}_t(y)$ and $\tl{q}_t(y)$ as the densities of $y$ when $x$ has distribution $p(x)$ and $q(x)$, respectively. Then, under some mild regularity conditions\footnote{See~\citet{lyu2009interpretation} for the details of regularity conditions}, we have
$$
\frac{d}{dt} D_{KL}( \tl{p}_t(y) \Vert \tl{q}_t(y)) = -\frac{1}{2}  D_F( \tl{p}_t(y) \Vert \tl{q}_t(y)).
$$
\end{lemma}
This provides us the interpretation that score matching (by minimizing Fisher divergence $D_F(p \Vert q_{\theta})$) looks for stability, where the optimal parameter $\theta$ leads to least changes in the KL divergence between the two models when a small amount of noise is added to the training data.

The above framework can be extended to matching the higher order score functions $\Pc_m(x)$ introduced in this paper, where the derivative is replaced by the $m$-th order derivative leading to minimizing\,\footnote{Subscript notation $\mathcal{L}$ is from \citet{lyu2009interpretation} where the Fisher divergence is generalized to any linear operator, e.g., higher order derivatives in our case.}
\begin{align*}
D_{\mathcal{L}}(p \Vert q_{\theta})=\int_{x} p(x) \left\| \frac{\nabla^{(m)}_x p(x)}{p(x)} - \frac{\nabla^{(m)}_x q_{\theta}(x)}{q_{\theta}(x)} \right\|^2 dx.
\end{align*}
Note that $\frac{\nabla^{(m)}_x p(x)}{p(x)} $ is exactly the $m$-th order score function $\Pc_m(x)$ up to sign. 


In addition,~\citet{swersky2011autoencoders} analyze the score matching for latent energy-based models with the joint distribution $p(x,h;\theta) =\frac{1}{Z(\theta)} \exp(-E_{\theta}(x,h))$, and provide the closed-form estimation for the parameters. Finally,~\citet{sasaki2014clustering} point out that the score function can be estimated efficiently through non-parametric methods without the need to estimate the density function. In fact, the solution is closed form, and the hyper-parameters (such as the kernel bandwidth and the regularization parameter) can be tuned easily through cross validation.

\paragraph{Estimation of the score function for $\phi(x)$:}
In some applications we need to  compute the score function of a nonlinear mapping of the input, i.e., for some function $\phi(x)$. This can be done by first estimating the joint density function of transformed variable and then computing its score function.
Let $t=\phi(x)$ and $D_t(i,j):=\left[ \frac{\partial x_i}{\partial t_j} \right]$. Then, we know
$$p_{\phi(x)}(t_1, \dotsc, t_r) = p_x(\phi_1^{-1}(t), \dotsc, \phi_r^{-1}(t)) \cdot \vert \det(D_t) \vert,$$
and the score function is defined as
$$\Pc_m(t) =(-1)^m \frac{\nabla^{(m)}_t p_{\phi(x)}(t)}{p_{\phi(x)}(t)}.$$


%


\subsection{Score function estimation in self-taught setting} \label{sec:score-self}

Now we discuss score function computation in the self-taught setting. 
Recall that in the self-taught setting, the distribution of unlabeled samples $\{\tilde{x}_i\}$ is different from the input of the labeled samples $\{x_i\}$. In Section~\ref{sec:problem-formulation},   we assume that the conditional distributions $p(\tilde{x}| \tilde{h})$ and $p(x|h)$, given the corresponding latent variables $\tilde{h}$ and $h$, are the same and give justifications for this assumption.

Under this assumption, estimating the score function for new samples $\{x_i\}$ is relatively straightforward, since we can transfer the estimated conditional distribution $p(\tl{x}|\tl{h})$ (using unlabeled samples $\{\tl{x}_i\}$) as the estimate for $p(x|h)$, and we can re-estimate the marginal distribution $p(h)$ easily. Thus, the use of  score functions allows for easy transfer of information under the self-taught framework with latent-variable modeling.

More concretely, for the estimation of higher order score function $\Pc_m(x)$, we need to estimate the joint probability density function of $x$ denoted  by $p(x)$. We have
\[
p(x) = \sum_h p(h) p(x|h) = \sum_h p(h) p(\tl{x}|h),
\]
where we also used the above assumption that the conditional distribution of target data $x$ given hidden variables can be substituted by the conditional distribution of unlabeled data $\tl{x}$ given hidden variables.
Note that $p(\tl{x}|h)$ can be estimated using unlabeled data $\{\tl{x}_i\}$. The unsupervised estimation of $p(\tl{x}|h)$ can be done in different ways, e.g., using spectral methods, score matching and so on.

%% file: theorems.tex
\section{Formal Statement of the Results} \label{sec:theorems}

In this section, we provide formal statement of the theorems characterizing the differential properties of the score functions. Before that, we propose an overview of notations mostly including tensor preliminaries.

\subsection{Notations and tensor preliminaries}\label{sec:notation}

Let $[n]$ denote the set $\{1,2,\dotsc,n\}$.

\paragraph{Tensor:} A real \emph{$r$-th order tensor} $T \in \bigotimes_{i=1}^r \R^{d_i}$ is a member of the outer product of Euclidean spaces $\R^{d_i}$, $i \in [r]$.
For convenience, we restrict to the case where $d_1 = d_2 = \dotsb = d_r = d$, and simply write $T \in \bigotimes^r \R^d$.
As is the case for vectors (where $r=1$) and matrices (where $r=2$), we may
identify a $r$-th order tensor with the $r$-way array of real numbers $[
T_{i_1,i_2,\dotsc,i_r} \colon i_1,i_2,\dotsc,i_r \in [d] ]$, where
$T_{i_1,i_2,\dotsc,i_r}$ is the $(i_1,i_2,\dotsc,i_r)$-th coordinate of $T$
with respect to a canonical basis. For convenience, we limit to third order tensors $(r=3)$ in our analysis, while the results for higher order tensors are also provided.


\paragraph{Tensor as multilinear form:} We view a tensor $T \in \Rbb^{d \times d \times d}$ as a multilinear form. 
Consider matrices $M_l \in \R^{d\times d_l}, l \in \{1,2,3\}$. Then tensor $T(M_1,M_2,M_3) \in \R^{d_1}\otimes \R^{d_2}\otimes \R^{d_3}$ is defined as
\begin{align} \label{eqn:multilinear form def}
T(M_1,M_2,M_3)_{i_1,i_2,i_3} := \sum_{j_1, j_2,j_3\in[d]} T_{j_1,j_2,j_3} \cdot M_1(j_1, i_1) \cdot M_2(j_2, i_2) \cdot M_3(j_3, i_3).
\end{align}
In particular, for vectors $u,v,w \in \R^d$, we have\,\footnote{Compare with the matrix case where for $M \in \R^{d \times d}$, we have $ M(I,u) = Mu := \sum_{j \in [d]} u_j M(:,j) \in \R^d$.}
\begin{equation} \label{eqn:rank-1 update}
 T(I,v,w) = \sum_{j,l \in [d]} v_j w_l T(:,j,l) \ \in \R^d,
\end{equation}
which is a multilinear combination of the tensor mode-$1$ fibers.
Similarly $T(u,v,w) \in \R$ is a multilinear combination of the tensor entries,  and $T(I, I, w) \in \R^{d \times d}$ is a linear combination of the tensor slices.

\paragraph{CP decomposition and tensor rank:} A $3$rd order tensor $T \in \Rbb^{d \times d \times d}$ is said to be rank-$1$ if it can be written in the form
\begin{align} \label{eqn:rank-1 tensor}
T= w \cdot a \otimes b\otimes c \Leftrightarrow T(i,j,l) = w \cdot a(i) \cdot b(j) \cdot c(l),
\end{align}
where notation $\otimes$  represents the {\em tensor (outer) product}, and $a \in \Rbb^d$, $b \in \Rbb^d$, $c \in \Rbb^d$ are unit vectors (without loss of generality).
A tensor $T  \in \Rbb^{d \times d \times d}$ is said to have a CP rank $k\geq 1$ if it can be written as the sum of $k$ rank-$1$ tensors
\begin{equation}\label{eqn:tensordecomp}
T = \sum_{i\in [k]} w_i a_i \otimes b_i \otimes c_i, \quad w_i \in \Rbb, \ a_i,b_i,c_i \in \Rbb^d.
\end{equation}


\paragraph{Derivative of tensor-valued functions:} Consider function $F(x) \in \bigotimes^r \R^d$ as a tensor-valued function with vector input $x \in \R^d$. The gradient of $F(x)$ w.r.t.\ variable $x$ is defined as a higher order tensor $\nabla_x F(x) \in \bigotimes^{r+1} \R^d$ such that
\begin{equation} \label{eqn:derivativedef}
\nabla_x F(x)_{i_1,\dotsc,i_r,j} := \frac{\partial F(x)_{i_1,\dotsc,i_r}}{\partial x_j}.
\end{equation}
In addition, the $m$-th order derivative is denoted by $\nabla_x^{(m)} F(x) \in \bigotimes^{r+m} \R^d$.

Finally, the transposition of a tensor with respect to a permutation matrix is defined as follows.
\begin{definition}[Tensor transposition]
Consider tensor $A \in \bigotimes^r \R^d$ and permutation vector $\pi = [\pi_1,\pi_2,\dotsc,\pi_r] \in \R^r$ as a permutation of index vector $1:r$. Then, the $\pi$-transpose of $A$ denoted by $A^{\langle \pi \rangle}$ is defined such that it satisfies
\[
A^{\langle \pi \rangle} (j_{\pi_1},\dotsc,j_{\pi_r}) = A(j_1,\dotsc,j_r).
\]
\end{definition}
In other words, the $i$-th mode of tensor $A^{\langle \pi \rangle}$ corresponds to the $\pi_i$-th mode of tensor $A$.

\subsection{Stein identity}

The following lemma states Stein's identity saying how first order score functions yield differential properties.

\begin{lemma}[Stein's lemma~\citep{stein2004use}] \label{steinslemma}
Let $x \in \mathbb{R}^{d_x}$ be a random vector with joint density function $p(x)$. Suppose the score function $\nabla_x \log p(x)$ exists. Consider any continuously differentiable tensor function $G(x):\mathbb{R}^{d_x} \rightarrow \bigotimes^r \mathbb{R}^{d_y}$ such that all the entries of $p(x) \cdot G(x)$ go to zero on the boundaries of support of $p(x)$. Then, we have
\begin{equation*} 
\Ebb[G(x) \otimes \nabla_x \log p(x)]=-\Ebb[\nabla_x G(x)],
\end{equation*}
Note that it is also assumed that the above expectations exist (in the sense that the corresponding integrals exist).
\end{lemma}
The proof follows integration by parts; the result for the scalar $x$ and scalar-output functions $g(x)$ is provided in~\citet{stein2004use}. 

\subsection{Parametric Stein identity} \label{sec:Stain_par}

%
We first recall and expand some parametric notations mentioned earlier.
Let $\Theta$ denote the set of parameters such that for $\theta \in \Theta$, $p(x;\theta)$ be a valid $\theta$-parametric probability density function. For $\theta_0 \in \Theta$, let $\mathcal{P}(\R^{d_x},\theta_0)$ be the collection of $\theta$-parametric probability density functions on $\mathbb{R}^{d_x}$ for which there exists a bounded neighborhood $\Theta_0 \subset \Theta$ of $\theta_0$ and an integrable function $l: \mathbb{R}^{d_x} \rightarrow \mathbb{R}^+$ such that $p(x;\theta) \leq l(x)$ over $\mathbb{R}^{d_x}$ for all $\theta \in \Theta_0$. Given $\theta_0 \in \Theta$ and $p \in \mathcal{P}(\R^{d_x},\theta_0)$, we write $x \sim p(\cdot; \theta_0)$ to denote the joint density function of $x$.

The following {\em regularity conditions} is defined along the lines of~\citep[Definition 2.1]{ley2013parametric}.

\begin{definition} \label{def_par}
Let $\theta_0$ be an interior point of $\Theta$ and $p \in \mathcal{P}(\R^{d_x},\theta_0)$. Define $S_{\theta} := \{x \in \R^{d_x} | p(x;\theta)>0\}$ as the support of $p(\cdot,\theta)$. We  define the class $\mathcal{G}(p, \theta_0)$ as the collection of functions $G: \mathbb{R}^{d_x} \times \Theta \rightarrow \bigotimes^r \mathbb{R}^{d_y} $ such that there exits $\Theta_0$ some neighborhood of $\theta_0$ where the following conditions are satisfied:
\begin{enumerate}
\item There exists a constant $c_g \in \mathbb{R}$ (not depending on $\theta$) such that $\int G(x;\theta)_{i_1,\dotsc,i_r} p(x; \theta) dx = c_g$ for all $\theta \in \Theta_0$. Note that the equality is entry-wise.
\item For all $x \in S_{\theta}$ the mapping $\theta \rightarrow G(\cdot;\theta)p(\cdot;\theta)$ is differentiable in the sense of distributions over $\Theta_0$, and in addition the order of derivative w.r.t.\ $\theta$ and integration over $x$ can be changed.
\end{enumerate}
\end{definition}

Finally, we state the parametric characterization of Stein's lemma as follows which is the result of \citet[Theorem 2.1]{ley2013parametric} generalized to tensor-output functions. 

\begin{theorem}[Parametric Stein characterization] \label{steinslemma_par}
Let $x \in \mathbb{R}^{d_x}$ be a random vector with joint $\theta$-parametric density function $p(x;\theta)$.
If the parametric score function $\nabla_{\theta} \log p(x;\theta)$ exists, then for all $G(x;\theta) \in \mathcal{G}(p;\theta_0)$ defined in Definition~\ref{def_par}, we have
\begin{equation*} 
\Ebb[G(x;\theta) \otimes \nabla_{\theta} \log p(x;\theta)]=-\Ebb[\nabla_{\theta}  G(x;\theta)] \quad \textnormal{at} \ \theta=\theta_0.
\end{equation*}
\end{theorem}

See Appendix~\ref{appendix:proof_higher} for the proof. The above result also holds for discrete random vectors; see~\citet[Section 2.4]{ley2013parametric} for the details.

\subsection{Higher order Stein identities}

We first provide the formal definition of higher order Score functions $\Pc_m(x)$, and then their differential properties are stated.

\begin{definition}[Higher order score functions] \label{def:diffoperator_high}
Let $p(x)$ denote the joint probability density function of random vector $x \in \R^d$.
We denote $\Pc_m(x) \in \bigotimes^m \R^d$ as the $m$-th order score function which is defined based on the recursive differential relation
\begin{equation} \label{eqn:diffoperator_recursion}
\Pc_m(x) := - \Pc_{m-1}(x) \otimes \nabla_x \log p(x) - \nabla_x \Pc_{m-1}(x),
\end{equation}
with $\Pc_0(x)=1$.
\end{definition}
By induction on $m$ we can prove that the above definition is equivalent to (see the proof in the appendix)
\begin{equation} \label{eqn:diffoperator}
\Pc_m(x)=(-1)^m \frac{\nabla_x^{(m)} p(x)}{p(x)}.
\end{equation}
Note that the first order score function $\Pc_1(x) = -\nabla_x \log p(x)$ is the same as score function in Stein's lemma; see Lemma~\ref{steinslemma}.
These higher order score functions enable us to generalize the Stein's identity in Lemma~\ref{steinslemma} to higher orders as follows.

\begin{theorem}[Yielding higher order differential operators] \label{thm:steins_higher}
Let $x \in \mathbb{R}^{d_x}$ be a random vector with joint density function $p(x)$. Suppose the $m$-th order score function $\Pc_m(x)$ defined in~\eqref{eqn:diffoperator_recursion} exists.
Consider any continuously differentiable tensor function $G(x):\mathbb{R}^{d_x} \rightarrow \bigotimes^r \mathbb{R}^{d_y}$ satisfying the regularity condition such that all the entries of $\nabla_x^{(i)} G(x) \otimes \Pc_{m-i-1}(x) \otimes p(x)$, $i \in \{0,1,\dotsc,m-1\}$, go to zero on the boundaries of support of $p(x)$. Then, we have
\[
\Ebb \left[ G(x) \otimes \Pc_m(x) \right] = \Ebb \left[ \nabla^{(m)}_x G(x) \right].
\]
\end{theorem}

The result can be proved by iteratively applying the recursion formula of score functions in~\eqref{eqn:diffoperator_recursion} and Stein's identity in Lemma~\ref{steinslemma}, see Appendix~\ref{appendix:proof_higher} for the details.

\subsection{Parametric higher order Stein identities}
Let\,\footnote{Here, $|\theta|$ denote  the  dimension of parameter $\theta$.} $\Pc_m(x;\theta) \in \bigotimes^m \R^{|\theta|}$ be the $m$-th order parametric score function which is defined based on the recursive differential relation
\begin{equation} \label{eqn:diffoperator_recursion_par}
\Pc_m(x;\theta) := - \Pc_{m-1}(x;\theta) \otimes \nabla_\theta \log p(x;\theta) - \nabla_\theta \Pc_{m-1}(x;\theta),
\end{equation}
with $\Pc_0(x;\theta)=1$.
By induction on $m$ we can prove that the above definition is equivalent to
\begin{align*}
\Pc_m(x;\theta)=(-1)^m \frac{\nabla_{\theta}^{(m)} p(x;\theta)}{p(x;\theta)}.
\end{align*}

\begin{theorem}[Yielding higher order parametric differential operators] \label{thm:steins_higher_par}
Let $x \in \mathbb{R}^{d_x}$ be a random vector with joint $\theta$-parametric density function $p(x;\theta)$.
If the $m$-th order parametric score function $\Pc_m(x;\theta)$ defined in~\eqref{eqn:diffoperator_recursion_par} exists, then for all $G(x;\theta) \in \mathcal{G}(p;\theta_0)$ defined in Definition~\ref{def_par}, we have
\[
\Ebb[G(x;\theta) \otimes \Pc_m(x;\theta) ] = \Ebb[\nabla_{\theta}^{(m)} G(x;\theta)] \quad \textnormal{at} \ \theta=\theta_0.
\]
\end{theorem}